\documentclass[11pt,a4paper]{article}
\usepackage{emnlp2018}
\usepackage[utf8]{inputenc}
\usepackage{amsmath}
\usepackage{booktabs}
\usepackage{multirow}
\usepackage{tikz}
\usetikzlibrary{automata, arrows, shapes}
\tikzset{every edge/.style={draw,->,>=stealth',shorten >=1pt,auto,anchor=south,sloped,semithick}}
\tikzset{initial text={},double distance=2pt}
\tikzset{state/.append style={fill=white,minimum size=6pt,inner sep=1pt}}

\usepackage{pgfplots}
\usepgfplotslibrary{groupplots}
\definecolor{s1}{RGB}{228, 26, 28}
\definecolor{s2}{RGB}{55, 126, 184}
\definecolor{s3}{RGB}{77, 175, 74}
\definecolor{s4}{RGB}{152, 78, 163}
\definecolor{s5}{RGB}{255, 127, 0}
\usepackage{txfonts}

\newcommand{\score}{s}

\newcommand{\eos}{\texttt{</s>}}
\renewcommand{\div}{\mathbin/}
\newcommand{\bigdiv}{\mathbin\bigg/}

\aclfinalcopy
\title{Correcting Length Bias in Neural Machine Translation}
\author{Kenton Murray \and David Chiang \\ Department of Computer Science and Engineering \\
University of Notre Dame \\ \texttt{\{kmurray4,dchiang\}@nd.edu}}

\begin{document}
\maketitle

\begin{abstract}
We study two problems in neural machine translation (NMT). First, in beam search, whereas a wider beam should in principle help translation, it often hurts NMT. Second, NMT has a tendency to produce translations that are too short. Here, we argue that these problems are closely related and both rooted in label bias. We show that correcting the brevity problem almost eliminates the beam problem; we 
compare some commonly-used methods for doing this, finding that a simple per-word reward works well; and we introduce a simple and quick way to tune this reward using the perceptron algorithm.
\end{abstract}

\section{Introduction}

Although highly successful, neural machine translation (NMT) systems continue to be plagued by a number of  problems. We focus on two here: the beam problem and the brevity problem.

First, machine translation systems rely on heuristics to search through the intractably large space of possible translations. Most commonly, beam search is used during the decoding process. Traditional statistical machine translation systems often rely on large beams to find good translations. However, in neural machine translation, increasing the beam size has been shown to degrade performance. 
This is the last of the six challenges identified by \citet{koehn+knowles:2017}. 

The second problem, noted by several authors, is that NMT tends to generate translations that are too short. \citeauthor{jean2015montreal} and
\citeauthor{koehn+knowles:2017}
address this by dividing translation scores by their length, inspired by work on audio chords \cite{boulanger2013audio}. A similar method is also used by Google's production system \cite{wu+al:2016}. A third simple method used by various authors \cite{och2002discriminative,he+al:2016,neubig2016lexicons} is a tunable reward added for each output word. \citet{huang2017finish} and \citet{yang2018breaking} propose variations of this reward that enable better guarantees during search.

In this paper, we argue that these two problems are related (as hinted at by \citeauthor{koehn+knowles:2017}) and that both stem from \emph{label bias}, an undesirable property of models that generate sentences word by word instead of all at once. 

The typical solution is to introduce a sentence-level correction to the model.
We show that making such a correction almost completely eliminates the beam problem. We compare two commonly-used corrections, length normalization and a word reward, and show that the word reward is slightly better.

Finally, instead of tuning the word reward using grid search, we introduce a way to learn it using a perceptron-like tuning method. We show that the optimal value is sensitive both to task and beam size, implying that it is important to tune for every model trained. Fortunately, tuning is a quick post-training step. 

\section{Problem}

Current neural machine translation models are examples of locally normalized models, which estimate the probability of generating an output sequence $e = e_{1:m}$ as
\[ P(e_{1:m}) = \prod_{i=1}^m P(e_i \mid e_{1:i-1}). \]

For any partial output sequence $e_{1:i}$, let us call $P(e' \mid e_{1:i})$, where $e'$ ranges over all possible completions of $e_{1:i}$, the \emph{suffix distribution} of $e_{1:i}$. The suffix distribution must sum to one, so if the model overestimates $P(e_{1:i})$, there is no way for the suffix distribution to downgrade it. This is known as \emph{label bias} 
\cite{bottou:1991,lafferty+al:2001}.

\subsection{Label bias in sequence labeling}

\begin{figure}
\begin{center}
\footnotesize
\begin{tikzpicture}[x=2.5cm,y=1cm]
\node[initial,state] at (0,0) (q0) {};
\node[state] at (1,1) (q1) {};
\draw (q0) edge node {$\text{a}/0.6$} (q1);
\node[state] at (2,2) (q3) {};
\draw (q1) edge node {$\text{helicopter}/0.6$} (q3);
\node[state] at (2,1) (q4) {};
\draw (q1) edge node[pos=0.7] {$\text{chopper}/0.3$} (q4);
\node[state] at (2,0) (q5) {};
\draw (q1) edge node [below] {$\text{whirlybird}/0.1$} (q5);

\node[state] at (1,-1) (q2) {};
\draw (q0) edge node[below] {$\text{an}/0.4$} (q2);
\node[state] at (2,-1) (q6) {};
\draw (q2) edge node[below] {$\text{autogyro}/1$} (q6);
\end{tikzpicture}
\end{center}
\caption{Label bias causes this toy word-by-word translation model to translate French \emph{un h\'elicopt\`ere} incorrectly to \emph{an autogyro}.}
\label{fig:labelbias}
\end{figure}
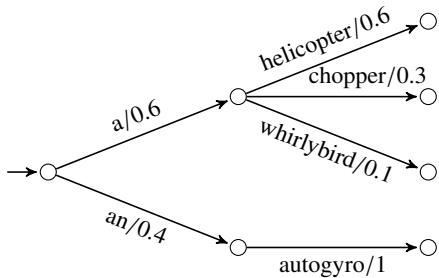

Label bias was originally identified in the context of HMMs and MEMMs for sequence-labeling tasks, where the input sequence $f$ and output sequence $e$ have the same length, and $P(e_{1:i})$ is conditioned only on the partial input sequence $f_{1:i}$. In this case, since $P(e_{1:i})$ has no knowledge of future inputs, it's much more likely to be incorrectly estimated. For example, suppose we had to translate, word-by-word, \emph{un h\'elicopt\`ere} to \emph{a helicopter} (Figure~\ref{fig:labelbias}). Given just the partial input \emph{un}, there is no way to know whether to translate it as \emph{a} or \emph{an}. Therefore, the probability for the incorrect translation $P(\text{an})$ will turn out to be an overestimate. As a result, the model will overweight translations beginning with \emph{an}, regardless of the next input word.

This effect is most noticeable when the suffix distribution has low entropy, because even when new input (\emph{h\'elicopt\`ere}) is revealed, the model will tend to ignore it. For example, suppose that the available translations for \emph{h\'elicopt\`ere} are \emph{helicopter}, \emph{chopper}, \emph{whirlybird}, and \emph{autogyro}. The partial translation \emph{a} must divide its probability mass among the three translations that start with a consonant, while \emph{an} gives all its probability mass to \emph{autogyro}, causing the incorrect translation \emph{an autogyro} to end up with the highest probability.

In this example, $P(\text{an})$, even though overestimated, is still lower than $P(\text{a})$, and wins only because its suffixes have higher probability. Greedy search would prune the incorrect prefix \emph{an} and yield the correct output. In general, then, we might expect greedy or beam search to alleviate some symptoms of label bias. Namely, a prefix with a low-entropy suffix distribution can be pruned if its probability is, even though overestimated, not among the highest probabilities. Such an observation was made by \citet{zhang+nivre:2012} in the context of dependency parsing, and we will see next that precisely such a situation affects output length in NMT.

\subsection{Length bias in NMT}

In NMT, unlike the word-by-word translation example in the previous section, each output symbol is conditioned on the entire input sequence. Nevertheless, it's still possible to overestimate or underestimate $p(e_{1:i})$, so the possibility of label bias still exists. We expect that it will be more visible with weaker models, that is, with less training data.

Moreover, in NMT, the output sequence is of variable length, and generation of the output sequence stops when \eos{} is generated. In effect, for any prefix ending with \eos{}, the suffix distribution has zero entropy. This situation parallels example of the previous section closely: if the model overestimates the probability of outputting \eos{}, it may proceed to ignore the rest of the input and generate a truncated translation. 

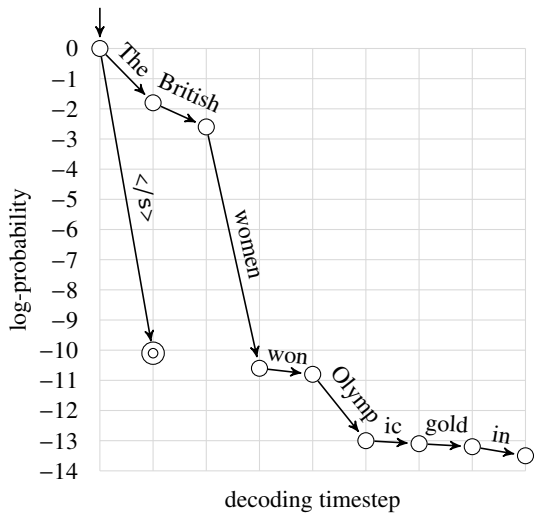
\begin{figure}
\begin{center} \small
\begin{tikzpicture}[x=0.7cm,y=0.4cm]
\node at (-1.5,-7) [rotate=90] {log-probability};
\node at (4,-15) {decoding timestep};
\foreach \y in {0,...,-14} {
  \draw [gray!30] (0,\y) -- (8,\y);
  \node[anchor=east] at (-0.25,\y) {$\y$};
}
\foreach \x in {0,...,8} {
  \draw [gray!30] (\x,0) -- (\x,-14);
}
\node [initial above,state] (q0) at (0,0) {};
\node [accepting,state] (qs) at (1, -10.1)  {}; \draw (q0) edge node {\eos} (qs);
\node [state] (q1) at (1, -1.8) {}; \draw (q0) edge node[pos=0.4] {The} (q1);
\node [state] (q2) at (2, -2.6) {}; \draw (q1) edge node {\strut British} (q2);
\node [state] (q3) at (3, -10.6) {}; \draw (q2) edge node {women} (q3);
\node [state] (q4) at (4, -10.8) {}; \draw (q3) edge node {won} (q4);
\node [state] (q5) at (5, -13.0) {}; \draw (q4) edge node {Olymp} (q5);
\node [state] (q6) at (6, -13.1) {}; \draw (q5) edge node {ic} (q6);
\node [state] (q7) at (7, -13.2) {}; \draw (q6) edge node {gold} (q7);
\node [state] (q8) at (8, -13.5) {}; \draw (q7) edge node {in} (q8);
\end{tikzpicture}
\end{center}
\caption{A locally normalized model must determine, at each time step, a ``budget'' for the total remaining log-probability. In this example sentence, ``The British women won Olymp ic gold in p airs row ing,'' the empty translation has initial position 622 in the beam. Already by the third step of decoding, the correct translation has a lower score than the empty translation. However, using greedy search, a nonempty translation would be returned.}
\label{fig:viz}
\end{figure}

Figure~\ref{fig:viz} illustrates how this can happen. Although the model can learn not to prefer shorter translations by predicting a low probability for $\eos$ early on, at each time step, the score of $\eos$ puts a limit on the total remaining score a translation can have; in the figure, the empty translation has score $-10.1$, so that no translation can have score lower than $-10.1$. This lays a heavy burden on the model to correctly guess the total score of the whole translation at the outset.

As in our label-bias example, greedy search would prune the incorrect empty translation. More generally, consider beam search: at time step $t$, only the top $k$ partial or complete translations are retained while the rest are pruned. (Implementations of beam search vary in the details, but this variant is simplest for the sake of argument.) Even if a translation ending at time $t$ scores higher than a longer translation, as long as it does not fall within the top $k$ when compared with partial translations of length $t$ (or complete translations of length at most $t$), it will be pruned and unable to block the longer translation. But if we widen the beam ($k$), then translation accuracy will suffer. We call this problem (which is \citeauthor{koehn+knowles:2017}'s sixth challenge) the \emph{beam} problem. Our claim, hinted at by \citet{koehn+knowles:2017}, is that the brevity problem and the beam problem are essentially the same, and that solving one will solve the other.

\section{Correcting Length}

To address the brevity problem, many designers of NMT systems add corrections to the model. These corrections are often presented as modifications to the search procedure. But, in our view, the brevity problem is essentially a modeling problem, and these corrections should be seen as modifications to the model (Section~\ref{sec:models}). Furthermore, since the root of the problem is local normalization, our view is that these modifications should be trained as globally-normalized models (Section~\ref{sec:training}).

\subsection{Models}
\label{sec:models}

Without any length correction, the standard model score (higher is better) is:
\[ \score(e) = \sum_{i=1}^m \log P(e_i \mid e_{1:i}). \]

To our knowledge, there are three methods in common use for adjusting the model to favor longer sentences.

\emph{Length normalization} divides the score by $m$ \cite{koehn+knowles:2017, jean2015montreal, boulanger2013audio}: 
\begin{equation*}
\score'(e) = \score(e) \div m.
\end{equation*}

\emph{Google's NMT system} \citep{wu+al:2016} relies on a more complicated correction:
\begin{align*}
\score'(e) &= \score(e) \bigdiv \frac{(5+m)^\alpha}{(5+1)^\alpha}.
\end{align*}

Finally, some systems add a constant \emph{word reward} \cite{he+al:2016}:
\begin{equation*}
\score'(e) = \score(e) + \gamma m.
\end{equation*}
If $\gamma=0$, this reduces to the baseline model.
The advantage of this simple reward is that it can be computed on partial translations, making it easier to integrate into beam search. 

\subsection{Training}
\label{sec:training}

All of the above modifications can be viewed as modifications to the base model so that it is no longer a locally-normalized probability model.

To train this model, in principle, we should use something like the globally-normalized negative log-likelihood:
\begin{align*}
L &= -\log \frac{\exp{s'(e^\ast)}}{\sum_{e} \exp{s'(e)}}
\end{align*}
where $e^\ast$ is the reference translation. However, optimizing this is expensive, as it requires performing inference on every training example or heuristic approximations \cite{andor+al:2016, shen2015minimum}.

Alternatively, we can adopt a two-tiered model, familiar from phrase-based translation \cite{och2002discriminative}, first training $s$ and then training $s'$ while keeping the parameters of $s$ fixed, possibly on a smaller dataset. A variety of methods, like minimum error rate training \cite{och:2003,he+al:2016}, are possible, but keeping with the globally-normalized negative log-likelihood, we obtain, 
for the constant word reward, the gradient:
\begin{align*}
\frac{\partial L}{\partial \gamma} &= -|e^\ast| + E[|e|].
\intertext{If we approximate the expectation using the mode of the distribution, we get}
\frac{\partial L}{\partial \gamma} &\approx -|e^\ast| + |\hat{e}|
\end{align*}
where $\hat{e}$ is the 1-best translation.
Then the stochastic gradient descent update is just the familiar perceptron rule:
\begin{align*}
\gamma &\leftarrow \gamma + \eta\,(|e^\ast| - |\hat{e}|),
\end{align*}
although below, we update on a batch of sentences rather than a single sentence.
Since there is only one parameter to train, we can train it on a relatively small dataset.

Length normalization does not have any additional parameters, with the result (in our opinion, strange) that a change is made to the model without any corresponding change to training. We could use gradient-based methods to tune the $\alpha$ in the GNMT correction, but the perceptron approximation turns out to drive $\alpha$ to $\infty$, so a different method would be needed.

\section{Experiments}\label{sec:experiments}

We compare the above methods in four settings, a high-resource German--English system, a medium-resource Russian--English system, and two low-resource French--English and English--French systems. For all settings, we show that larger beams lead to large BLEU and METEOR drops if not corrected. We also show that the optimal parameters can depend on the task, language pair, training data size, as well as the beam size. These values can affect performance strongly.

\subsection{Data and settings}

Most of the experimental settings below follow the recommendations of \citet{denkowski-neubig:2017:NMT}. Our high-resource, German--English data is from the 2016 WMT shared task \cite{bojar2016findings}. We use a bidirectional encoder-decoder model with attention \cite{bahdanau+al:2015}.\footnote{We use Lamtram \cite{neubig15lamtram} for all experiments and our modifications have been added to the project.} Our word representation layer has 512 hidden units, while other hidden layers have 1024 nodes. Our model is trained using Adam with a learning rate of 0.0002. We use 32k byte-pair encoding (BPE) operations learned on the combined source and target training data \cite{sennrich2015neural}. We train on minibatches of size 2012 words and validate every 100k sentences, selecting the final model based on development perplexity. 

Our medium-resource, Russian--English system uses data from the 2017 WMT translation task, which consists of roughly 1 million training sentences \cite{bojar2017findings}. 
We use the same architecture as our German--English system, but only have 512 nodes in all layers. We use 16k BPE operations and dropout of 0.2. We train on minibatches of 512 words and validate every 50k sentences.

Our low-resource systems use French and English data from the 2010 IWSLT TALK shared task \cite{paul2010overview}. We build both French--English and English--French systems. These networks are the same as for the medium Russian-English task, but use only 6k BPE operations. We train on minibatches of 512 words and validate every 30k sentences, restarting Adam when the development perplexity goes up.

To tune our correction parameters, we use 1000 sentences from the German--English development dataset, 1000 sentences from the Russian--English development dataset, and the entire development dataset for French--English (892 sentences)\footnote{We found through preliminary experiments that this size of dev subset was an adequate trade-off between tuning speed and performance.}. We initialize the parameter, $\gamma=0.2$. 
We use batch gradient descent, which we found to be much more stable than stochastic gradient descent, and use a learning rate of $\eta = 0.2$, clipping gradients 
for $\gamma$ 
to 0.5.  Training stops if all parameters have an update of less than 0.03 or a max of 25 epochs was reached.


\begin{table*}
\centering\begin{tabular}{ll|rrrrrr}

\toprule
\multicolumn{2}{c|}{Russian--English (medium)} & \multicolumn{6}{c}{Beam Size} \\
& & 10 & 50 & 75 & 100 & 150 & 1000 \\

\midrule
baseline & BLEU & 24.9 & 23.8 & 23.6 & 23.3 & 22.5 & 3.7  \\
& METEOR & 30.9 & 30.0 & 29.7 & 29.4 & 28.8 & 12.8 \\
& length & 0.90 & 0.86 & 0.85 & 0.84 & 0.81  & 0.31   \\
\midrule
reward & BLEU & 26.5 & 26.6 & 26.5 & 26.5 & 26.5 & 25.7   \\
& METEOR & 32.0 & 32.0 & 31.9 & 31.9 & 31.9 & 31.2 \\
& length & 0.98 & 0.98 & 0.98 & 0.98 & 0.98 & 1.02    \\
& $\gamma$ & 0.716 & 0.643 & 0.640 & 0.633 & 0.617 & 0.562 \\
\midrule
norm & BLEU & 26.2 & 26.3 & 26.3 & 26.3 & 26.3 & 25.3 \\
& METEOR & 31.8 & 31.8 & 31.8 & 31.7 & 31.7 & 31.2 \\
& length & 0.96 & 0.96 & 0.96 & 0.96 & 0.97 & 1.02 \\

\bottomrule
\end{tabular}
\caption{Results of the Russian--English translation system. We report BLEU and METEOR scores, as well as the ratio of the length of generated sentences compared to the correct translations (length). $\gamma$ is the word reward score discovered during training. Here, we examine a much larger beam (1000). The beam problem is more pronounced at this scale, with the baseline system losing over 20 BLEU points when increasing the beam from size 10 to 1000. However, both our tuned length reward score and length normalization recover most of this loss.}
\label{tab:russian}
\end{table*}

\begin{table*}
\centering

\begin{tabular}{ll|rrr}
\toprule
 \multicolumn{2}{c|}{German--English (large)} &  \multicolumn{3}{c}{Beam Size} \\
&  &  10 & 50 & 75  \\
\midrule
baseline & BLEU &  29.6 & 28.6 & 28.2 \\
& METEOR & 34.0 & 33.1 & 32.8 \\
& length & 0.95 & 0.90 & 0.89  \\
\midrule
reward & BLEU & 30.3 & 30.6 & 30.6 \\
& METEOR & 34.9 & 34.8 & 34.9  \\
& length &  1.02 & 1.00 & 1.00 \\
& $\gamma$ & 0.67 & 0.57 & 0.58 \\
\midrule
norm & BLEU & 30.7 & 31.0 & 30.9 \\
& METEOR & 34.9 & 35.0 & 35.0 \\
& length & 1.00 & 1.00 & 1.00 \\
\bottomrule
\end{tabular}

\caption{Results of the high-resource German--English system. Rows: BLEU, METEOR, length = ratio of output to reference length; $\gamma$ 
= learned parameter value.  
While baseline performance decreases with beam size due to the brevity problem, other methods perform more consistently across beam sizes. Length normalization (norm) gets the best BLEU scores, but similar METEOR scores to the word reward.}
\label{tab:german}
\end{table*}

\begin{table*}
\centering

\begin{tabular}{ll|rrrrr}
\toprule
\multicolumn{2}{c|}{French--English (small)} & \multicolumn{5}{c}{Beam Size}  \\
& & 10 & 50 & 100 & 150 & 200  \\
\midrule
baseline & BLEU & 30.0 & 28.9 & 25.4 & 21.9 & 19.4 \\
& METEOR & 32.4 & 31.3 & 28.6 & 25.9 & 24.1  \\
& length & 0.94 & 0.89 & 0.80 & 0.71  & 0.64  \\
\midrule
reward & BLEU & 29.4 & 29.7 & 29.7 & 29.8 & 29.8 \\
& METEOR & 32.8 & 32.9 & 32.9 & 32.9 & 32.9 \\
& length & 1.03 & 1.03 & 1.03 & 1.03  & 1.03 \\
& $\gamma$ & 1.20 & 1.05 & 1.01 & 0.99 & 0.97 \\
\midrule
norm & BLEU & 30.7 & 30.8 & 30.7 & 30.7 & 30.7 \\
& METEOR & 32.8 & 32.8 & 32.8 & 32.7 & 32.7 \\
& length & 0.97 & 0.97 & 0.97 & 0.96 & 0.96 \\
\bottomrule
\\[2ex]
\toprule
\multicolumn{2}{c|}{English--French (small)}& \multicolumn{5}{c}{Beam Size}  \\
&  & 10 & 50 & 100 & 150 & 200  \\
\midrule
baseline & BLEU & 25.8 & 26.1 & 26.1 & 25.5 & 24.3 \\
& METEOR & 47.8 & 47.5 & 47.2 & 46.3 & 44.2  \\
& length & 1.03 & 1.01 & 1.00 & 0.97  & 0.92  \\
\midrule
reward & BLEU & 25.5 & 25.5 & 25.5 & 25.5 & 25.5 \\
& METEOR & 48.3 & 48.5 & 48.5 & 48.5 & 48.4  \\
& length & 1.05 & 1.05 & 1.05 & 1.05  & 1.05 \\
& $\gamma$ & 0.353 & 0.444 &0.465 & 0.474 & 0.475 \\
\midrule
norm & BLEU & 25.4 & 25.5 & 25.5 & 25.5 & 25.5 \\
& METEOR & 48.4 & 48.4 & 48.4 & 48.4 & 48.4   \\
& length & 1.06 & 1.05 & 1.05 & 1.05 & 1.05 \\

\bottomrule
\end{tabular}
\caption{Results of low-resource French--English and English--French systems. Rows: BLEU, METEOR, length = ratio of output to reference length; $\gamma$ 
= learned parameter value. 
While baseline performance decreases with beam size due to the brevity problem, other methods perform more consistently across beam sizes. Word reward gets the best scores in both directions on METEOR. Length normalization (norm) gets the best BLEU scores in Fra-Eng due to the slight bias of BLEU towards shorter translations.}
\label{tab:french}
\end{table*}

\begin{table*}
\centering\begin{tabular}{l|rrrrrr}
\toprule
beam & 10 & 50 & 75 & 100 & 150 & 200 \\ 
\midrule
French--English (small) & 6.9 & 27.2 & 52.4 & 71.1 & 105.9 & 176.6 \\
English--French (small) & 12.6 & 44.2 & 67.3 & 88.1 & 107.5 & 111.2 \\
German--English (large) & 6.8 & 132.6 & 1066  \\
\bottomrule
\end{tabular}
\caption{Tuning time on top of baseline training time. Times are in minutes on 1000 dev examples (German--English) or 892 dev examples (French--English). Due to the much larger model size, we only looked at beam sizes up to 75 for German--English.}
\label{tab:times}
\end{table*}

\subsection{Solving the length problem solves the beam problem}

Here, we first show that the beam problem is indeed the brevity problem. We then demonstrate that solving the length problem does solve the beam problem. 
Tables~\ref{tab:russian}, \ref{tab:german}, and \ref{tab:french} show the results of our German--English, Russian--English, and French--English systems respectively. Each table looks at the impact on BLEU, METEOR, and the ratio of the lengths of generated sentences compared to the gold lengths \cite{papineni2002bleu, denkowski:lavie:meteor-wmt:2014}. The baseline method is a standard model without any length correction. The reward method is the tuned constant word reward discussed in the previous section. Norm refers to the normalization method, where a hypothesis' score is divided by its length.

\subsubsection{Baseline}

The top sections of Tables~\ref{tab:russian}, \ref{tab:german}, \ref{tab:french} illustrate the brevity and beam problems in the baseline models. As beam size increases, the BLEU and METEOR scores drop significantly. This is due to the brevity problem, which is illustrated by the length ratio numbers that also drop with increased beam size. For larger beam sizes, the length of the generated output sentences are a fraction of the lengths of the correct translations. For the lower-resource French--English task, the drop is more than 8 BLEU when increasing the beam size from 10 to 150. The issue is even more evident in our Russian-English system where we increase the beam to 1000 and BLEU scores drop by more than 20 points. 

\subsubsection{Word reward}

The results of tuning the word reward, $\gamma$, as described in Section \ref{sec:training}, is shown in the second section of Tables~\ref{tab:russian}, \ref{tab:german}, and \ref{tab:french}. In contrast to our baseline systems, our tuned word reward always fixes the brevity problem (length ratios are approximately 1.0), and generally fixes the beam problem. An optimized word reward score always leads to improvements in METEOR scores over any of the best baselines. Across all language pairs, reward and norm have close METEOR scores, though the reward method wins out slightly. BLEU scores for reward and norm also increase over the baseline in most cases, despite BLEU's inherent bias towards shorter sentences. 
Most notably, whereas the baseline Russian--English system lost more than 20 BLEU points when the beam was increased to 1000, our tuned reward score resulted in a BLEU gain over any baseline beam size. Whereas in our baseline systems, the length ratio decreases with larger beam sizes, our tuned word reward results in length ratios of nearly 1.0 across all language pairs, mitigating many of the issues of the brevity problem.

\subsubsection{Wider beam}

We note that the beam problem in NMT exists for relatively small beam sizes -- especially when compared to traditional beam sizes in SMT systems. On our medium-resource Russian--English system, we investigate the full impact of this problem using a much larger beam size of 1000. In Table \ref{tab:russian}, we can see that the beam problem is particularly pronounced. The first row of the table shows the uncorrected, baseline score. From a beam of 10 to a beam of 1000, the drop in BLEU scores is over 20 points. This is largely due to the brevity problem discussed earlier. The second row of the table shows the length of the translated outputs compared to the lengths of the correct translations. Though the problem persists even at a beam size of 10, at a beam size of 1000, our baseline system generates less than one third the number of words that are in the correct translations. Furthermore, 37.3\% of our translated outputs have sentences of length 0. In other words, the most likely translation is to immediately generate the stop symbol. This is the problem visualized in Figure \ref{fig:viz}.

However, when we tune our word reward score with a beam of 1000, the problem mostly goes away. Over the uncorrected baseline, we see a 22.0 BLEU point difference for a beam of 1000. Over the uncorrected baseline with a beam of 10, the corrected beam of 1000 gets a BLEU gain of 0.8 BLEU. However, the beam of 1000 still sees a drop of less than 1.0 BLEU over the best corrected version. The word reward method beats the uncorrected baseline and the length normalization correction in almost all cases.

\begin{figure*}
\begin{center}
\includegraphics[width=0.7\textwidth]{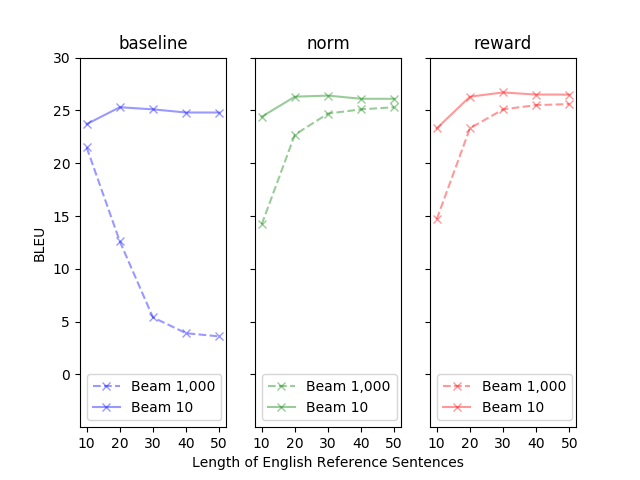}
\caption{Impact of beam size on BLEU score when varying reference sentence lengths (in words) for Russian--English. The x-axis is cumulative moving right; length 20 includes sentences of length 0-20, while length 10 includes 0-10. As reference length increases, the BLEU scores of a baseline system with beam size of 10 remain nearly constant. However, a baseline system with beam 1000 has a high BLEU score for shorter sentences, but a very low score when the entire test set is used. Our tuned reward and normalized models do not suffer from this problem on the entire test set, but take a slight performance hit on the shortest sentences.}
\label{fig:short}
\end{center}
\end{figure*}

\begin{figure*}
    \centering
    \includegraphics[width=0.8\textwidth]{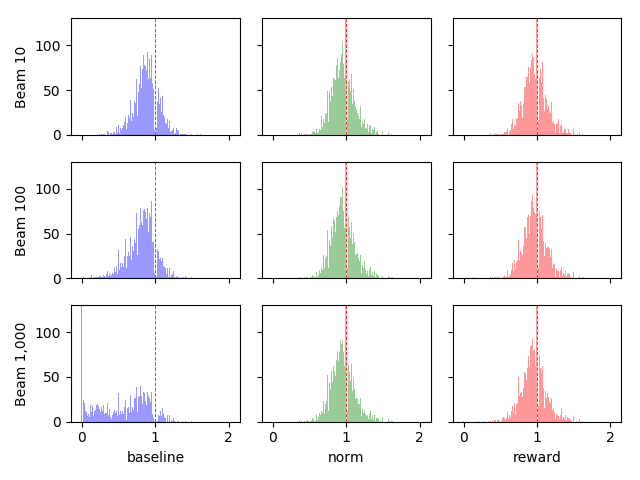}
    \caption{Histogram of length ratio between generated sentences and gold varied across methods and beam size for Russian--English. Note that the baseline method skews closer 0 as the beam size increases, while our other methods remain peaked around 1.0. There are a few outliers to the right that have been cut off, as well as the peaks at 0.0 and 1.0.}
    \label{fig:all_histogram}
\end{figure*}

\subsubsection{Short sentences}

Another way to demonstrate that the beam problem is the same as the brevity problem is to look at the translations generated by baseline systems on shorter sentences. Figure \ref{fig:short} shows the BLEU scores of the Russian--English system for beams of size 10 and 1000 on sentences of varying lengths, with and without correcting lengths. The x-axes of the figure are cumulative: length 20 includes sentences of length 0--20, while length 10 includes 0--10. It is worth noting that BLEU is a word-level metric, but the systems were built using BPE; so the sequences actually generated are longer than the x-axes would suggest.

The baseline system on sentences with 10 words or less still has relatively high BLEU scores---even for a beam of 1000. Though there is a slight drop in BLEU (less than 2), it is not nearly as severe as when looking at the entire test set (more than 20). When correcting for length with normalization or word reward, the problem nearly disappears when considering the entire test set, with reward doing slightly better. For comparison, the rightmost points in each of the subplots correspond to the BLEU scores in columns 10 and 1000 of Table \ref{tab:russian}. This suggests that the beam problem is strongly related to the brevity problem.

\subsubsection{Length ratio}

The interaction between the length problem and the beam problem can be visualized in the histograms of Figure \ref{fig:all_histogram} on the Russian--English system. In the upper left plot, the uncorrected model with beam 10 has the majority of the generated sentences with a length ratio close to 1.0, the gold lengths. Going down the column, as the beam size increases, the distribution of length ratios skews closer to 0. By a beam size of 1000, 37\% of the sentences have a length of 0. However, both the word reward and the normalized models remain very peaked around a length ratio of 1.0 even as the beam size increases.

\subsection{Tuning word reward}

Above, we have shown that fixing the length problem with a word reward score fixes the beam problem. However these results are contingent upon choosing an adequate word reward score, which we have done in our experiments by optimization using a perceptron loss. Here, we show the sensitivity of systems to the value of this penalty, as well as the fact that there is not one correct penalty for all tasks. It is dependent on a myriad of factors including, beam size, dataset, and language pair.

\subsubsection{Sensitivity to $\gamma$}

\begin{figure}

\centering \small
\begin{tabular}{cc}
\begin{tikzpicture}
\begin{groupplot}[group style = {group size = 1 by 2,vertical sep=0.5cm},width=7.5cm,ylabel near ticks]
\nextgroupplot[ylabel={BLEU},ymin=20,ymax=26,height=3.5cm,xmin=0,xmax=1.2,xtick={0,0.2,...,1.2},xticklabels={}, title={Russian--English}]
\addplot coordinates { (0.0,21.6) (0.2,22.5) (0.4,23.7) (0.6,24.6) (0.8,24.7) (1.0,22.9) (1.2,21.0) };
\addplot[mark=none] coordinates { (0.643,0) (0.643,100) };
\nextgroupplot[ylabel={length ratio},
xmin=0,xmax=1.2,ymin=0.8,ymax=1.3,height=3.5cm]
\addplot coordinates { (0.0,0.829) (0.2,0.856) (0.4,0.897) (0.6,0.949) (0.8,0.998) (1.0,1.067) (1.2,1.162) };
\addplot[mark=none] coordinates { (0.643,0) (0.643,2) };
\addplot[mark=none] { 1 };
\end{groupplot}
\end{tikzpicture}
\end{tabular}

\centering \small
\begin{tabular}{cc}
\begin{tikzpicture}
\begin{groupplot}[group style = {group size = 1 by 2,vertical sep=0.5cm},width=7.5cm,ylabel near ticks]
\nextgroupplot[ylabel={BLEU},ymin=18,ymax=24,height=3.5cm,xmin=0,xmax=1.2,xtick={0,0.2,...,1.2},xticklabels={}, title={German--English}]
\addplot coordinates { (0.0,22.5) (0.2,23.1) (0.4,23.6) (0.6,23.0) (0.8,22.0) (1.0,21.1) (1.2,18.4) };
\addplot[mark=none] coordinates { (0.572,0) (0.572,100) };
\nextgroupplot[ylabel={length ratio},xlabel={word reward ($\gamma$)}, xmin=0,xmax=1.2,ymin=0.9,ymax=1.4,height=3.5cm]
\addplot coordinates { (0.0,0.929) (0.2,0.960) (0.4,0.996) (0.6,1.043) (0.8,1.103) (1.0,1.162) (1.2,1.317) };
\addplot[mark=none] coordinates { (0.572,0) (0.572,2) };
\addplot[mark=none] { 1 };
\end{groupplot}
\end{tikzpicture}
\end{tabular}

\caption{Effect of word penalty on BLEU and hypothesis length for Russian--English (top) and German-English (bottom) on 1000 unseen dev examples with beams of 50. Note that the vertical bars represent the word reward that was found during tuning.}
    \label{fig:untuned}

\end{figure}
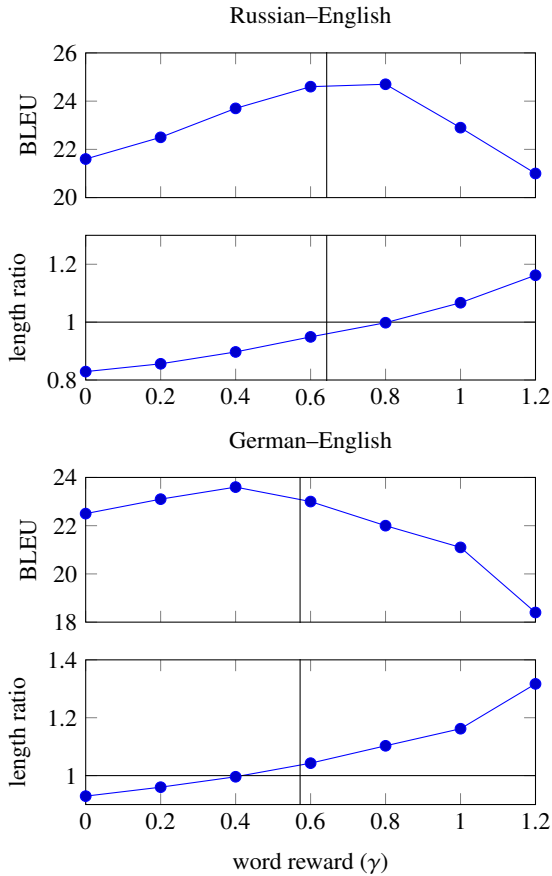

In order to investigate how sensitive a system is to the reward score, we varied values of $\gamma$ from $0$ to $1.2$ on both our German--English and Russian--English systems with a beam size of 50. BLEU scores and length ratios on 1000 heldout development sentences are shown in Figure \ref{fig:untuned}. The length ratio is correlated with the word reward as expected, and the BLEU score varies by more than 5 points for German--English and over 4.5 points for Russian--English. On German--English, our method found a value of $\gamma=0.57$, which is slightly higher than optimal; this is because the heldout sentences have a slightly shorter length ratio than the training sentences. Conversely, on Russian--English, our found value of $\gamma=0.64$ is slightly lower than optimal as these heldout sentences have a slightly higher length ratio than the sentences used in training.

\subsubsection{Optimized $\gamma$ values}

Tuning the reward penalty using the method described in Section \ref{sec:training} resulted in consistent improvements in METEOR scores and length ratios across all of our systems and language pairs. Tables~\ref{tab:russian}, \ref{tab:german}, and \ref{tab:french} show the optimized value of $\gamma$ for each beam size. Within a language pair, the optimal value of $\gamma$ is different for every beam size. Likewise, for a given beam size, the optimal value is different for every system. Our French--English and English--French systems in Table \ref{tab:french} have the exact same architecture, data, and training criteria. Yet, even for the same beam size, the tuned word reward scores are very different.

\paragraph{Training dataset size}

Low-resource neural machine translation performs significantly worse than high-resource machine translation \cite{koehn+knowles:2017}. Table \ref{tab:varying} looks at the impact of training data size on BLEU scores and the beam problem by using 10\% and 50\% of the available Russian--English data. Once again, the optimal value of $\gamma$ is different across all systems and beam sizes. Interestingly, as the amount of training data decreases, the gains in BLEU using a tuned reward penalty increase with larger beam sizes. This suggests that the beam problem is more prevalent in lower-resource settings, likely due to the fact that less training data can increase the effects of label bias.

\begin{table*}
\centering\begin{tabular}{rr|rrrrr}
\toprule
\multicolumn{2}{c|}{Russian--English (medium)} & \multicolumn{5}{c}{Beam Size} \\
Dataset Size & & 10 & 50 & 75 & 100 & 150 \\

\midrule
 & baseline & 24.9 & 23.8 & 23.6 & 23.3 & 22.5  \\
100\%& reward & 26.5 & 26.6 & 26.5 & 26.5 & 26.5   \\
& $\gamma$ & 0.716 & 0.643 & 0.640 & 0.633 & 0.617  \\
\midrule
 & baseline & 22.8 & 21.4  & 20.8 & 20.4 & 19.2   \\
50\% & reward & 24.7 & 25.0 & 24.9 & 24.9 & 25.0 \\
& $\gamma$ & 0.697 & 0.645 & 0.638 & 0.636 & 0.646 \\
\midrule
 & baseline & 17.0 & 16.2  & 15.8 & 15.6 & 15.1   \\
10\% & reward & 17.6 & 18.0 & 18.0 & 18.0 & 18.1 \\
& $\gamma$ & 0.892 & 0.835 & 0.773 & 0.750 & 0.800 \\

\bottomrule
\end{tabular}
\caption{Varying the size of the Russian--English training dataset results in different optimal word reward scores ($\gamma$). In all settings, the tuned score alleviates the beam problem. As the datasets get smaller, using a tuned larger beam improves the BLEU score over a smaller tuned beam. This suggests that lower-resource systems are more susceptible to the beam problem.}
\label{tab:varying}
\end{table*}

\subsubsection{Tuning time}

Fortunately, the tuning process is very inexpensive. Although it requires decoding on a development dataset multiple times, we only need a small dataset. The time required for tuning our French--English and German--English systems is shown in Table \ref{tab:times}. These experiments were run on an Nvidia GeForce GTX 1080Ti. The tuning usually takes a few minutes to hours, which is just a fraction of the overall training time. We note that there are numerous optimizations that could be taken to speed this up even more, such as storing the decoding lattice for partial reuse. However, we leave this for future work.

\subsection{Word reward vs. length normalization}

Tuning the word reward score generally had higher METEOR scores than length normalization across all of our settings. With BLEU, length normalization beat the word reward on German-English and French--English, but tied on English-French and lost on Russian--English. For the largest beam of 1000, the tuned word reward had a higher BLEU than length normalization. Overall, the two methods have relatively similar performance, but the tuned word reward has the more theoretically justified, globally-normalized derivation -- especially in the context of label bias' influence on the brevity problem.

\section{Conclusion}

We have explored simple and effective ways to alleviate or eliminate the beam problem. We showed that the beam problem can largely be explained by the brevity problem, which results from the locally-normalized structure of the model. We compared two corrections to the model and introduced a method to learn the parameters of these corrections. Because this method is helpful and easy, we hope to see it included to make stronger baseline NMT systems.

We have argued that the brevity problem is an example of label bias, and that the solution is a very limited form of globally-normalized model. These can be seen as the simplest case of the more general problem of label bias and the more general solution of globally-normalized models for NMT \cite{wiseman2016sequence, venkatraman2015improving, ranzato2015sequence, shen2015minimum}. Some questions for future research are:
\begin{itemize}
\item Solving the brevity problem leads to significant BLEU gains; how much, if any, improvement remains to be gained by solving label bias in general?
\item Our solution to the brevity problem requires globally-normalized training on only a small dataset; can more general globally-normalized models be trained in a similarly inexpensive way?
\end{itemize}

\section*{Acknowledgements}

This research was supported in part by University of Southern California, subcontract 67108176 under DARPA contract HR0011-15-C-0115, and an Amazon Research Award to Chiang.


\bibliographystyle{acl_natbib_nourl}
\bibliography{references}
\end{document}